\pdfoutput=1

\documentclass[11pt]{article}

\usepackage{acl}

\usepackage{times}
\usepackage{latexsym}
\usepackage{graphicx}
\usepackage{color,soul}
\usepackage{hyperref}
\usepackage{amsmath}
\usepackage{subfigure}
\usepackage{hyperref}
\usepackage{booktabs}
\usepackage{algorithm}
\usepackage{algorithmic}
\usepackage{circledsteps}
\usepackage{multirow}

\usepackage[shortlabels]{enumitem}
\usepackage[font=small,skip=1pt]{caption}
\usepackage{colortbl}

\makeatletter
    \newcommand*{\algrule}[1][\algorithmicindent]{\makebox[#1][l]{\hspace*{.5em}\thealgruleextra\vrule height \thealgruleheight depth \thealgruledepth}}%
\newcommand*{\thealgruleextra}{}
\newcommand*{\thealgruleheight}{.75\baselineskip}
\newcommand*{\thealgruledepth}{.25\baselineskip}

\newcount\ALG@printindent@tempcnta
\def\ALG@printindent{%
    \ifnum \theALG@nested>0
        \ifx\ALG@text\ALG@x@notext
        \else
            \unskip
            \addvspace{-1pt}
            \ALG@printindent@tempcnta=1
            \loop
                \algrule[\csname ALG@ind@\the\ALG@printindent@tempcnta\endcsname]%
                \advance \ALG@printindent@tempcnta 1
            \ifnum \ALG@printindent@tempcnta<\numexpr\theALG@nested+1\relax
            \repeat
        \fi
    \fi
    }%
\usepackage{etoolbox}
\patchcmd{\ALG@doentity}{\noindent\hskip\ALG@tlm}{\ALG@printindent}{}{
}
\makeatother

\newbox\statebox
\newcommand{\myState}[1]{%
    \setbox\statebox=\vbox{#1}%
    \edef\thealgruleheight{\dimexpr \the\ht\statebox+1pt\relax}%
    \edef\thealgruledepth{\dimexpr \the\dp\statebox+1pt\relax}%
    \ifdim\thealgruleheight<.75\baselineskip
        \def\thealgruleheight{\dimexpr .75\baselineskip+1pt\relax}%
    \fi
    \ifdim\thealgruledepth<.25\baselineskip
        \def\thealgruledepth{\dimexpr .25\baselineskip+1pt\relax}%
    \fi
    \State #1%
    \def\thealgruleheight{\dimexpr .75\baselineskip+1pt\relax}%
    \def\thealgruledepth{\dimexpr .25\baselineskip+1pt\relax}%
}

\usepackage[T1]{fontenc}

\usepackage[utf8]{inputenc}

\usepackage{microtype}

\usepackage{inconsolata}
\usepackage{authblk}

\title{Entropy Guided Extrapolative Decoding to Improve Factuality in Large Language Models}

\author[1]{\textbf{Souvik Das} \thanks{Work done during an internship at Tencent AI Lab, Bellevue, WA. Correspondence: \texttt{souvikda@buffalo.edu}}} 
\author[2]{\textbf{Lifeng Jin}} 
\author[2]{\textbf{Linfeng Song}} 
\author[2]{\textbf{Haitao Mi}} 
\author[2]{\textbf{Baolin Peng}} 
\author[2]{\textbf{Dong Yu}} 
\affil[1]{Department of Computer Science and Engineering, University at Buffalo, NY. }
\affil[2]{Tencent AI Lab, Bellevue, WA}

\begin{document}
\maketitle
\begin{abstract}
Large language models (LLMs) exhibit impressive natural language capabilities but suffer from hallucination -- generating content ungrounded in the realities of training data. Recent work has focused on decoding techniques to improve factuality during inference by leveraging LLMs’ hierarchical representation of factual knowledge, manipulating the predicted distributions at inference time. Current state-of-the-art approaches refine decoding by contrasting early-exit distributions from a lower layer with the final layer to exploit information related to factuality within the model forward procedure. However, such methods often assume the final layer is the most reliable and the lower layer selection process depends on it. In this work, we first propose extrapolation of critical token probabilities beyond the last layer for more accurate contrasting. We additionally employ layer-wise entropy-guided lower layer selection, decoupling the selection process from the final layer. Experiments demonstrate strong performance - surpassing state-of-the-art on multiple different datasets by large margins. Analyses show different kinds of prompts respond to different selection strategies. Our source code will be available in GitHub.\footnote{\href{https://github.com/souvikdgp16/extrapolative_decoding}{https://github.com/souvikdgp16/extrapolative\_decoding}}
\end{abstract}

\section{Introduction}

Despite their impressive capabilities \cite{brown2020language, openai2023gpt4} in natural language tasks, large language models (LLMs) tend to hallucinate -- generating content that does not align with real-world facts they were exposed to during pretraining \cite{Ji_2023} -- which poses deployment challenges \cite{guerreiro2023hallucinations}. 
The propensity of large language models for fabricating content remains an issue under active investigation. Overcoming hallucination is thus a significant challenge for safe and trustworthy AI applications, which becomes ever more important as their abilities expand through scaling. 

\begin{figure}[t]
\centering
\includegraphics[width=0.4\textwidth]{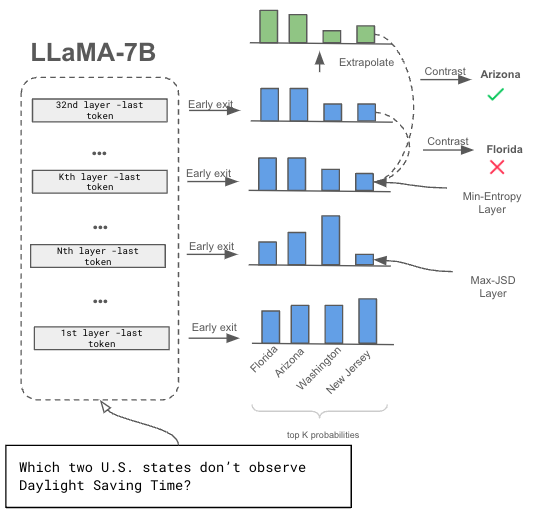}
\caption{Our proposed extrapolative decoding, final transformer layer is extrapolated to a predetermined layer before contrasting with a lower layer.}
\label{fig:extrapol_overview}
\end{figure}

Causes of hallucination may stem from flaws permeating the entire pipeline,  such as inaccurate, biased data, lack of grounding and consistency guardrails and suboptimal knowledge integration~\cite{li-etal-2022-pre, liška2022streamingqa, chang-etal-2019-bias, yin-etal-2023-large} . Promising avenues involve enforcing factual fidelity in generation~\cite{shi2023replug}, causal reasoning capacities~\cite{kıcıman2023causal}, and transparent, controllable knowledge deployment to temper fabrication~\cite{touvron2023llama}. 
Recently efforts have been focusing on inference techniques that improve factuality.
\citet{chuang2023dola} leverage the hierarchical factual knowledge encoded within LLMs, with lower layers capturing surface patterns and higher ones more semantic information. Inspired by \citet{li-etal-2023-contrastive}, they introduce DoLa - a strategy refining factual decoding by dynamically selecting and contrasting logits from lower or \textit{premature}  layers with the final or \textit{mature} layer. By exploiting the change in distributions from a lower and less contextualized layer to the last and most contextualized layer, DoLa showcases the potential for reducing hallucinations through utilizing the distribution \textit{maturation} process through the layers. Despite the success of this decoding strategy,  the method relies on the high maturity level of the last layer, which may not be true. Additionally,  the selection of the less mature layer is dependent on the final layer, which assumes that the most immature layer is the one furthest away from the last layer. This dependency on the last layer may not be desirable, especially when the last layer is not mature. 

The final predicted distribution can be made more mature by adding more transformer layers, which essentially extends the depth of the model. However, this is impractical because the extension may be dynamic and therefore expensive. In this work, we first propose inference-time \textit{logit extrapolation} to address this issue. Specifically, we extrapolate probabilities of specific tokens increasing or decreasing monotonically over the last few transformer layers, which enables the predicted distribution to become even more mature. Furthermore, we exploit the correlation between uncertainty-based metrics like entropy and factuality, i.e., tokens comprising factual sentences tend to exhibit higher probability and lower entropy. In contrast, tokens resulting in hallucinations generally originate from flatter distributions with greater uncertainty. Based on this observation, we exploit layer-wise token entropy as the selection criterion to select the lower contrasting layer that would lead to a better contrastive objective. In this way, we remove the dependency on the final layer from the selection process, which could alleviate the cascading effect of generating a factually false answer when using a premature final layer for guidance. 


Figure \ref{fig:extrapol_overview} shows an example of our method. The final layer's predictions is both incorrect in its prediction and premature in layer selection, where the model is insufficiently confident about the correct answer "\textit{Arizona}". Contrasting such uncertain distributions with lower layers can then erroneously produce inaccurate outputs like "\textit{Florida}". However, allowing critical token probabilities to continue evolve by extrapolation provides greater maturity to higher layers. More peaked, confident predictions in turn enable targeted contrasting to selectively refine immature lower-level tendencies, without overriding correct distributions. Thus, by avoiding preemptive interference and allowing further development of predictive maturity, our method generates factual responses like "\textit{Arizona}". Additionally, our entropy-based lower layer selection mitigates the dependency on final layer. This demonstrated case highlights this advantage, where entropy identifies the appropriate lower layer regardless of how inaccurate the final distribution is.

Our approach demonstrates strong performance on tasks related to factuality, outperforming the baseline methods by large margins on a variety of factuality-related tasks, such as TruthfulQA \cite{lin2022truthfulqa} and FACTOR\cite{factor}. Experiments further exhibit benefits for factual reasoning, with higher performance on StrategyQA\cite{strategyqa} and GSM8K\cite{gsm8k}. These gains highlight the broad efficacy of our method for not just isolated to factual recall but complex reasoning chains dependent on accurate intermediate deductions. Our evaluation validates the proposed approach as an promising inference-time decoding method for mitigating hallucination and enhancing truthfulness.



\section{Preliminaries}
\begin{figure*}
  \centering
    \includegraphics[width=0.75\textwidth]{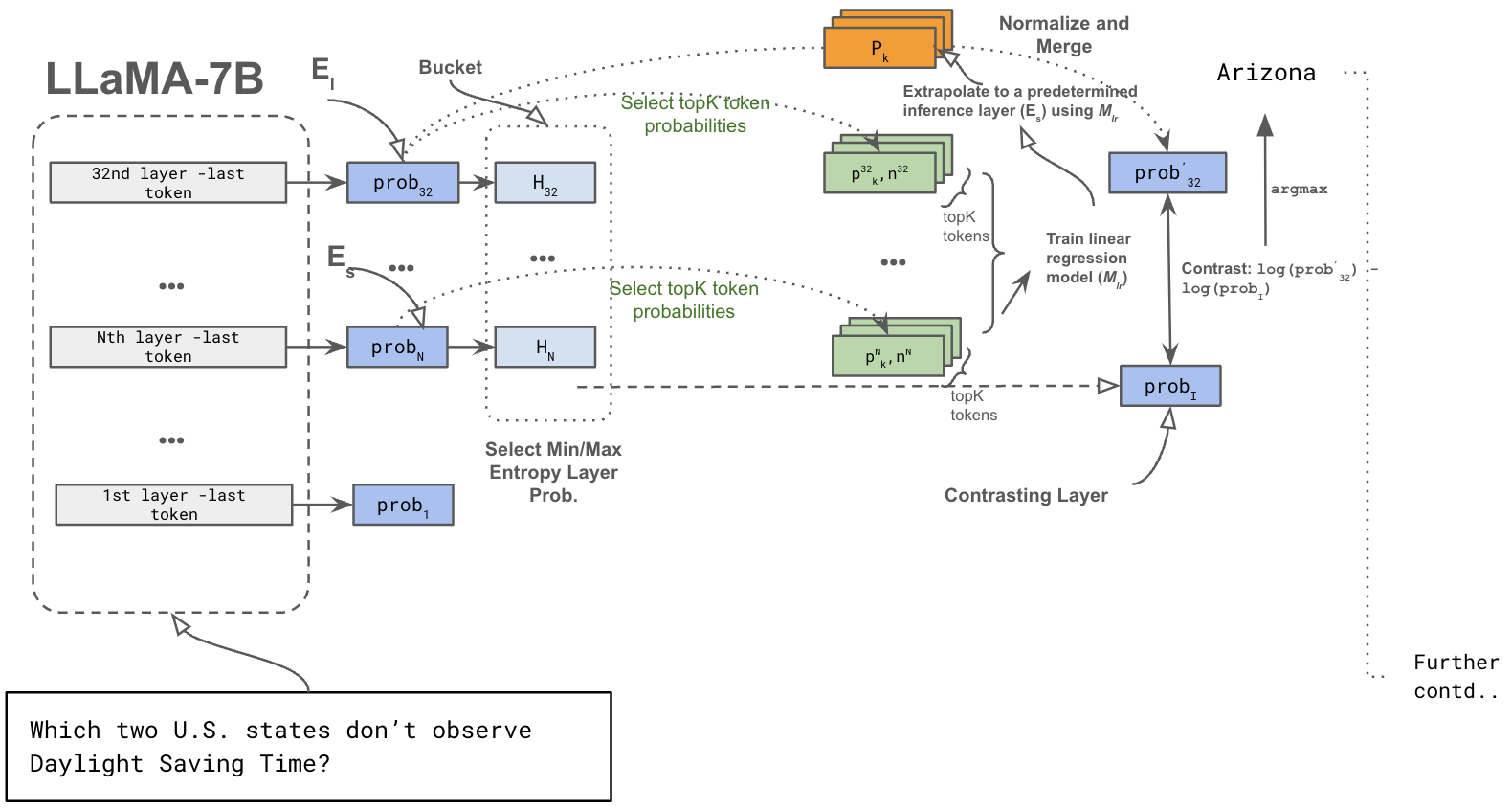}
    \caption{Overview of our entire inference pipeline. }
    \label{fig:whole_schema}
\end{figure*}

\subsection{Contrastive Decoding and Factuality}

Large language models usually have an embedding layer and $N$ stacked layers, and also an affine layer $\phi(.,.)$ to predict the probability of the next token.  Given a sequence of tokens $\mathrm{x}_p = \{x_1 ... x_{t-1}\}$, embedding layer first processes the tokens into sequence of vectors $\mathbf{h}_0=\{h_1^{(0)} ... h_{t-1}^{(0)}\}$, subsequently $\mathbf{h}_0$ would be processed by each of the transformer layers, where the output of $j$-th layer is denoted as $\mathbf{h}_j$. Then, the linear vocabulary head $\phi(.,.)$ predicts the probability of the next token $x_t$:

{
\begin{equation}
    p(x_t|x_{< t}) = \mathrm{softmax}(\phi(h_t^N)_t)
\end{equation}
}

Where $x_t$ $\in$ $\mathcal{V}$, the vocabulary set. Recently, \citet{chuang2023dola} has proposed a contrastive decoding \cite{li-etal-2023-contrastive} method, where instead of using an amateur model, they are contrasting the most \textit{mature layer} $N$ with a \textit{premature layer} $j$. The contrastive objective is defined as:

{
\begin{equation}
    \mathcal{L}_{CD} = \mathrm{log}p(x_t|x_{< t}) - \mathrm{log}q(x_t|x_{< t}) 
\end{equation}
}

Where $q(x_t|x_{< t})=\mathrm{softmax}(\phi(h_t^j)_t)$ is the probability of generating the next token derived from a lower transformer layer, i.e., $j<N$ which is also known as \textit{early-exit}. The \textit{premature layer} $j$ is selected by a dynamic selection metric $d(.,.)$, the Jensen-Shannon divergence between the mature layer and the candidate premature layers. The premature layer with the highest JSD is then selected as the appropriate premature layer within a predefined bucket of transformer layers $\mathcal{K}$, such as the 2nd bucket containing 10 layers from the 11th to the 20th layer $(10, 20]$.

\section{Methodology}



\subsection{Dynamic Contrasting Layer Selection}

To maximize the effect of contrastive decoding, we dynamically select a contrasting layer 
based on the entropy of the distribution from early-exit
 within a range of transformer layers. Mathematically, token-wise entropy can be represented as:
{
\begin{equation}
    \mathcal{H}_{ij} = - \sum_{x_t\in \mathcal{V}}p_{ij}(.|x_{<t})\mathrm{log}p_{ij}(.|x_{<t})
\end{equation}
}
where $p_{ij}(.|x_{<t})$ is the probability of the word being generated at the $j$-th token of the $i$-th transformer layer. We utilize both maximum entropy and minimum entropy as our selection strategies. The most optimal contrasting layer $\mathcal{I}$ is selected in this fashion:
{
\begin{equation}
    \mathcal{I} = 
    \begin{cases}
        \mathrm{arg}\mathop{min}_{i\in \mathcal{K}}(\mathcal{H}_{ij}) & \text{if } Q \in Q_s\\
        \mathrm{arg}\mathop{max}_{i\in \mathcal{K}}(\mathcal{H}_{ij})             & \text{otherwise} ,
    \end{cases}
\end{equation}
}
where $Q$ is the prompt, $Q_s$ is the set of open-ended prompts (more details in \S \ref{sec:entropy_analysis}), $\mathcal{K}$ is the range of transformer layers, which serves as a search space for the most optimal contrasting layer. For LLaMA-based models, following \citet{chuang2023dola}, we divide the transformer layers into 2-4 buckets based on model size to limit our search space to some specific layers.

\subsection{Logit Extrapolation}
Previous methods assume the last layer is the most mature. However, it might be possible that the assumed mature layer has room for more growth. Generally, it is very challenging to get a more mature representation without adding more transformer layers. We propose a very simple yet effective strategy to extrapolate the probabilities of a few critical tokens by extrapolating the probabilities using linear regression, shown in Algorithm \ref{alg:algorithm_extrap}. We consider the model’s last $3$ layers, and the extrapolation process is triggered only when the entropy in the last layer is changed drastically compared to the previous two layers.\footnote{This is determined by JS Distance, as explained in Algorithm \ref{alg:algorithm_extrap}}
\begin{algorithm}[H]
\caption{Logits Extrapolation}
\label{alg:algorithm_extrap}
\scriptsize
\textbf{Input:} Last $\mathcal{L}$ hidden layers of transformer for the last token $H_{1..\mathcal{L}}$, extrapolation trigger threshold $\alpha$, top k $t_k$ value, extrapolation start layer $E_s$, extrapolation end layer $E_l$ and extrapolation inference  layer $E_i$\\
\textbf{Output:} Extrapolated last layer probabilities: $\mathrm{prob_{\mathcal{L}}}'$ , if needed
\begin{algorithmic}[1] 
        \STATE $\mathrm{prob_{1..\mathcal{L}}} \gets \mathrm{softmax}(\phi(H_{1..\mathcal{L}}))$ \COMMENT{$\phi(.)$ is feed-forward network}
        \IF{$||\frac{\mathrm{JSD}(\mathrm{prob}_{\mathcal{L}}, \mathrm{prob}_{\mathcal{L}-1}) - \mathrm{JSD}(\mathrm{prob}_{\mathcal{L}-1}, \mathrm{prob}_{\mathcal{L}-2})}{\mathrm{JSD}(\mathrm{prob}_{\mathcal{L}-1}, \mathrm{prob}_{\mathcal{L}-2})}|| > \alpha$}
            \STATE for $t_k$ and  $\mathrm{prob_{1..\mathcal{L}}}$ starting from layer $E_s$ and ending at $E_l$, get layer-wise top k tokens probability: $p_k$ $\leftarrow$
              $\mathrm{top\_k(prob}_{E_s .. E_l})$
            \FOR{$i \gets 1$ to $t_k$}
                \IF{$\mathrm{is\_monotonic}(p_{k_i})$}
                    \STATE \textbf{continue}
                \ELSE
                    \STATE remove $p_{k_i}$
                \ENDIF
            \ENDFOR
    
            \STATE train a linear regression model $\mathcal{M}_{lr}$ using $p_{k}$ and layer numbers from $E_s$ to  $E_l$ \COMMENT{\textbf{Ref.} \S \ref{sec:lr_details}}
            \STATE get extrapolated probabilities $P_k \gets \mathcal{M}_{lr}(E_i)$
            \STATE $\mathrm{Normalize\_TopK}(P_k,p_k)$ to make sure top k probabilities remain as top k.
            \STATE $\mathrm{prob_{\mathcal{L}}}'$ $\gets$ $\mathrm{merge}(P_k, \mathrm{prob_{\mathcal{L}}})$
            \RETURN{} $\mathrm{prob_{\mathcal{L}}}'$
        \ENDIF
        \RETURN{} $\mathrm{prob_{\mathcal{L}}}$
\end{algorithmic}
\end{algorithm}
The extrapolation process begins with gathering probabilities of top k $t_k$ tokens from layer $E_s$ and ends at layer $E_l$. Then, we check whether the probabilities are monotonically increasing or decreasing from $E_s$ to $E_l$. We only keep the tokens where this monotonicity criterion is met. Then a linear regression model $\mathcal{M}_{lr}$ is trained using the collected probabilities(More details in \S \ref{sec:lr_details}). Using $\mathcal{M}_{lr}$, we extrapolate the probabilities to a predetermined inference layer $E_i$. The extrapolated probabilities are normalized such that the probabilities are still the highest in the distribution, but with potential change in their ranking. The normalization process is as follows:
{
\begin{align}
    & \mathrm{Normalize\_TopK}(P_k, p_k)_i \nonumber \\
    &= 
    \begin{cases}
        p_{k_i}, & \text{if } \mathrm{index}(P_{k_i}) \notin \mathrm{top\_k}\\
        P_{k_i},              & \text{otherwise}
    \end{cases}
\end{align}
}
Here, $p_k$ is the probabilities of top k tokens and $P_k$ is the corresponding extrapolated probability. Finally, we merge the extrapolated top k probabilities with the original probabilities.

\subsection{Training Linear Regression Model} \label{sec:lr_details}

The primary objective is to learn a regression model $\mathcal{M}_{lr}$ using the probabilities of top k($t_k$) vocabulary tokens $p_k$ starting from extrapolation start layer $E_s$ to extrapolation end layer $E_l$. For the extrapolation model in every time step, the training data is a pair of the layer number $n^j$(for example, in the range of $[0-32]$ for LLaMA-7B) and the corresponding token probability $p^j_{k_i}$ for a particular layer. To summarize we have the following training data: $[(n^{E_s}, p^{E_s}_{k_{i}}),..,(n^j,p^{j}_{k_i}),.., (n^{E_l}, p^{E_l}_{k_{i}})]^{t_k}_{i=0}$. We train and infer the regression model in batch size of $t_k$. During inference the extrapolated probabilities of each token is obtained by passing the predetermined inference layer $E_i$. More details in \S \ref{sec:lr_details}.

\subsection{Contrastive Objective}

Given the optimal contrasting($\mathcal{I}$) and mature layers obtained, we aim to amplify the output from the mature layer by further extrapolating critical token probabilities while downplaying the output from the contrasting layer. Following the Contrastive Decoding approach from \cite{li-etal-2023-contrastive}, we subtract the log probabilities of the contrasting layer outputs from those of the inflection layer. We define contrastive objective $\mathcal{L}_{CD}$, using which we get the final probabilities for decoding as:

{
\small
\begin{equation}
    \mathcal{L}_{CD} = 
\begin{cases}
    \mathrm{log}\frac{\mathrm{Extrapolate}(p(x_t|x_{< t}))}{q_{\mathcal{I}}(x_t|x_{< t})},& \text{if } x_t\in \mathcal{C}_a(x_t|x<t)\\
    - \infty,              & \text{otherwise}
\end{cases}
\end{equation}
}

\noindent Here, $p(x_t|x_{< t})$,  $q_{\mathcal{I}}(x_t|x_{< t})$ are the probability distributions of the mature and contrasting layers. $\mathrm{Extrapolate}(.)$ method calls Algorithm \ref{alg:algorithm_extrap}. We also incorporate the same \textit{adaptive plausibility constraint} strategy as in \cite{li-etal-2023-contrastive}. Here $\mathcal{C}_a(x_t|x<t)$ is a subset of $\mathcal{V}$ which signifies the output token probabilities are high enough from the mature layer:

{
\small
\begin{equation}
    \mathcal{C}_a(x_t|x<t) = \{x_t \in \mathcal{V}: p(x_t|x_{< t}) \leq \beta \mathop{\mathrm{max}}_w(p(w|x_{< t}))\}
\end{equation}
}

Here, $\beta$ is a hyperparameter in $[0, 1] $ that truncates the next token distribution in the mature layer. More details in \S \ref{sec:inf_details}.

\section{Experimentation}
\begin{figure*}
  \centering
    \includegraphics[width=0.9\textwidth]{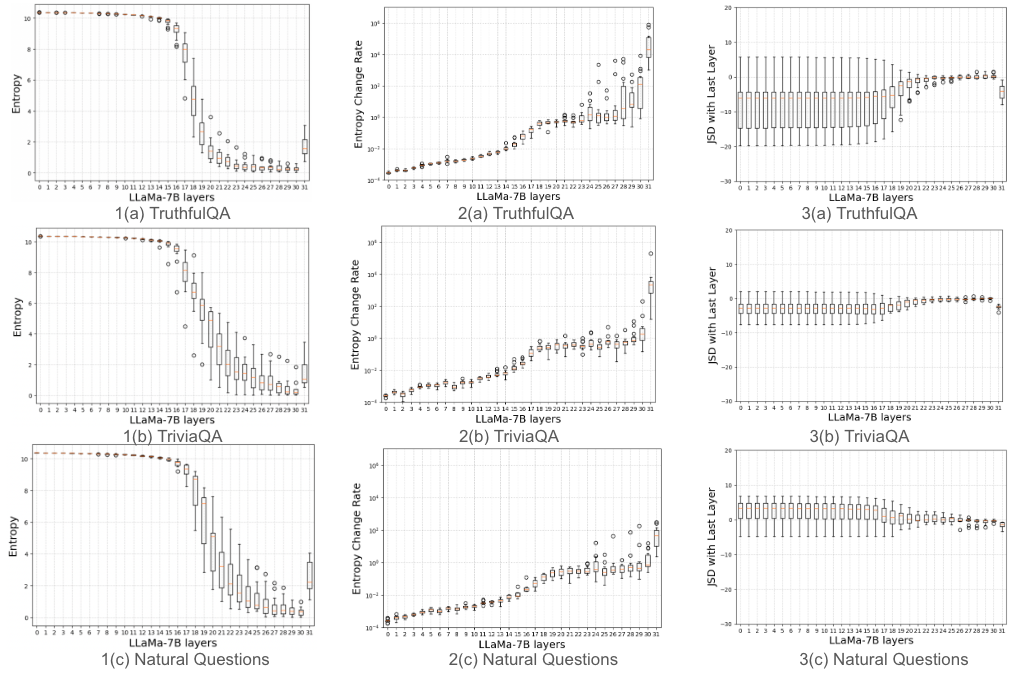}
    \caption{Analysis performed on $100$ prompts sampled from TruthfulQA, TriviaQA and Natural Questions. We plot three sets of graphs (1) Entropy($\mathcal{H}_i$) v/s Transformer layers (2) Entropy change rate i.e. $\delta(\mathcal{H}_i,\mathcal{H}_{i-1})/\mathcal{H}_{i-1}$ v/s Transformer layers (3) JSD with last layer v/s Transformer layers.} 
    \label{fig:graphs}
\end{figure*}
\subsection{Tasks}
 We consider two types of tasks for this work: the first is \textit{multiple choice} and the second one is \textit{open-ended generation} task. For the first task, we use the TruthfulQA dataset’s multiple choice split and the FACTOR dataset’s wiki split. We use the log probabilities of the choices to calculate a score and then make the choice.  For the second task, we consider the TruthfulQA dataset’s generation split. The answers were rated by GPT3 fine-tuned models for \textit{truthfulness} and \textit{informativeness}, and the evaluation process strictly follows previous procedures mentioned in the TruthfulQA paper. Furthermore, we use StrategyQA and GSM8K datasets. These datasets require chain-of-thought reasoning. If the generated answer contains the correct keywords, we consider it to be correct.

\subsection{Baselines}

\begin{itemize}[noitemsep,nolistsep, leftmargin=*]
\item \textbf{Original decoding}: we use greedy decoding.

\item \textbf{Inference Time Intervention (ITI)}\cite{li2023inferencetime}: ITI uses LLaMA-7B and a linear classifier trained on TruthfulQA to identify a set of heads that exhibit superior linear probing accuracy for answering factual questions.

\item \textbf{Contrastive Decoding (CD)}:  we follow the contrastive decoding setup proposed by \cite{chuang2023dola}, with LLaMA 7B as the amateur model and subsequent higher parameter models as expert models. For LLaMa 7B, we skipped the contrastive decoding results.

\item \textbf{DoLa}: this baseline uses a contrastive decoding strategy where a lower layer selected dynamically, instead of an amateur model, is used as the contrasting layer. 

\end{itemize}
\begin{figure}
    \centering
    \includegraphics[width=0.9\linewidth]{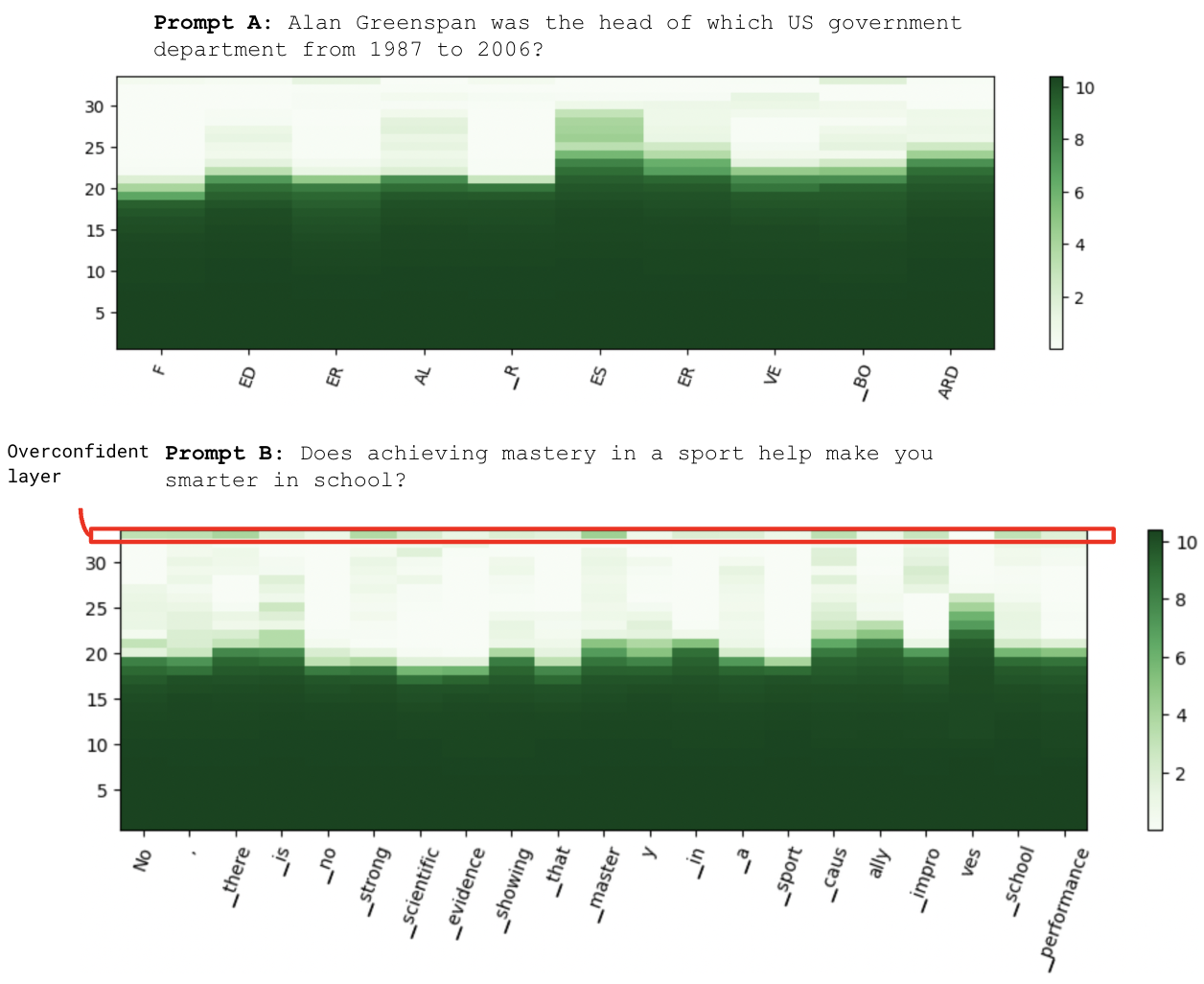}
    \caption{\textbf{Prompt A}: An example of factual prompt $Q_f$ and layer-wise entropy for LLaMA 7B. \textbf{Prompt B}: An example of open-ended prompt $Q_s$ and layer-wise entropy for LLaMA 7B, with annotated higher overconfident layer(more details in \S \ref{sec:entropy_analysis}), where there is a sudden increase in entropy. }
    \label{fig:enter-label}
\end{figure}
\subsection{Entropy Across Transformer Layers} \label{sec:entropy_analysis}

There is a correlation between uncertainty-based metrics like entropy $\mathcal{H}$ and model factuality \citeauthor{manakul2023selfcheckgpt}. Factual sentences are likely to contain tokens with higher likelihood and lower entropy, while hallucinations will likely come from positions with flat
probability distributions with high uncertainty. However, in this work, we observe different behaviors from two kinds of prompts: (1) factual prompts denoted as $Q_f$ where there is solely information needed like this: \textit{Alan Greenspan was the head of which US government department from 1987 to 2006?} They are found in datasets like TriviaQA, Natural Questions(NQ), etc. (2) Open-ended prompts denoted as $Q_s$ where the answer may not be found in commonly used training data. Prompts like \textit{Does achieving mastery in a sport help make you smarter in school?} can be found in TruthfulQA dataset. We analyzed these prompt categories by sampling $100$ prompts from TruthfulQA, TriviaQA, and NQ \footnote{We used TriviaQA and NQ for analysis as is completely factual in nature and prompts are of short length(average words: $16$). However, we did not use these datasets in evaluations due to large number of data-points in test split and lack of previous baselines. More details can be found in \S \ref{sec:reason_data}} and observing their entropy changes through layers of LLaMA 7B . Each prompt is a concatenation of question and answer: \texttt{<Question> <Answer>}, and we use the probabilities of only the answer tokens in our downstream analysis. As shown in Figure \ref{fig:graphs}, we plotted three metrics with the transformer layers: (1) Entropy, (2) Entropy change rate, and (3) JSD with the last transformer layer. The following observations were made:
\begin{itemize}[noitemsep,nolistsep, leftmargin=*]
    \item[$-$] Though the entropy is highest in lower layers for all the models, the entropy fluctuation for the TruthfulQA dataset is considerably low in the higher layers compared to TriviaQA and NQ. This suggests the higher layers in TruthfulQA are more overconfident, i.e. the model becomes confident about the prediction, however, in fact it is wrong. This is further bolstered by the fact an \textit{early-exit} performance in layers preceding final layer is considerably lower than the final layer.\footnote{Early-exit layer 28, LLaMA 7B, MC1: 21.3}
    \item[$-$] Entropy change rate is higher in higher layers in TruthfulQA, which suggests that the model constantly changes its predictions over the last few sequence of transformer layers. Meanwhile, for the other datasets, the slow change suggests that the model has been decided early.
    \item[$-$] In the third set of graphs, the spread of JSD between the last layer and other layers is high in TruthfulQA for the lower layers; this again suggests that lower layers are far more premature than the factual dataset's lower layers. Thus more likely it will be close to embedding layer where the contrast benefit is low.
\end{itemize}

Based on this analysis, we hypothesize that for open-ended prompts (like ones in TruthfulQA), the layers will be more premature than factual prompts, thereby suggesting the contrasting layer, after which the probabilities start to move in the truthful direction will lie in the higher layers with minimum entropy and vice versa for factual datasets (like TriviaQA and the other datasets in evaluation).

\subsection{Setup}

We use LLaMA series (7B, 13B, 33B, and 65B) models for all our experiments. The $0$-th layer corresponds to the word embedding layer before the first transformer layer. We divide the layers of LLaMA 7/13/33/65B models into 2/4/4/4 buckets of candidate layers. The hyperparameter search used 2-4 validation runs depending on the model. We do 2-fold validation for all the data sets to select the optimal buckets. For the TruthfulQA dataset, we assume all the prompts are of type $Q_s$(open-ended) and use minimum entropy configuration to select the contrasting layer. For other datasets, we assume all the prompts are of type $Q_f$(factual) and use maximum entropy configuration. More details can be found in \S \ref{sec:inf_details} along with hyperparameters in Table \ref{tab:tfqa_hyp}, \ref{tab:tfqa_hyp_2}.
\vspace{-5pt}
\section{Results}
\subsection{Multiple Choice}
For TruthfulQA multiple choice split, we adopt the same prompting strategy proposed by \citet{lin2022truthfulqa}. We use a minimum entropy setting for this dataset, and for all the models, the highest buckets are selected after 2-fold validation. Table \ref{tab:truthfulqa_baseline } shows significant performance improvement for LLaMA models in four sizes, outperforming the state-of-the-art baseline DoLa. 

The FACTOR(wiki) multiple choice dataset has a long paragraph as context with an answer and three distractor options. We use the maximum entropy setting for this dataset as most of the queries are factual; for all the models, the lowest buckets are selected after 2-fold validation. As evident from Table \ref{tab:truthfulqa_baseline }, our method outperforms DoLa.

\begin{table}[]
\centering
\scalebox{0.5}{%
\begin{tabular}{@{}lrrr|r@{}}
\toprule
\multicolumn{4}{c|}{\textbf{TruthfulQA-MC}}                                                                                                          & \multicolumn{1}{l}{\textbf{FACTOR-Wiki}} \\ \midrule
\multicolumn{1}{l|}{\textbf{Model/Method}} & \multicolumn{1}{l}{\textbf{MC1}($\uparrow$)} & \multicolumn{1}{l}{\textbf{MC2}($\uparrow$)} & \multicolumn{1}{l|}{\textbf{MC3}($\uparrow$)} & \multicolumn{1}{l}{\textbf{Accuracy}($\uparrow$)}    \\ \midrule
\multicolumn{1}{l|}{LLaMA7B}               & 25.6                             & 40.6                             & 19.2                              & 58.6                                     \\
\multicolumn{1}{l|}{LLaMA7B+ITI}           & 25.9                             & \multicolumn{1}{l}{-}            & \multicolumn{1}{l|}{-}            & \multicolumn{1}{l}{-}                    \\
\multicolumn{1}{l|}{LLaMA7B+DoLa}          & 32.2                             & \textbf{63.8}                    & 32.1                              & 62.2                                     \\
\multicolumn{1}{l|}{LLaMA7B+Ours}          & \textbf{36.1}                    & 63.7                             & \textbf{37.0}                    & \textbf{63.1}                            \\ \midrule
\multicolumn{1}{l|}{LLaMA13B}              & 28.3                             & 43.3                             & 20.8                              & 62.6                                     \\
\multicolumn{1}{l|}{LLaMA13B+CD}           & 24.4                             & 41.0                               & 19.0                                & 64.4                                     \\
\multicolumn{1}{l|}{LLaMA13B+DoLa}         & 28.9                             & 64.9                             & 34.8                              & 66.2                                     \\
\multicolumn{1}{l|}{LLaMA13B+Ours}         & \textbf{32.1}                    & \textbf{67.0}                      & \textbf{37.9}                    & \textbf{66.7}                            \\ \midrule
\multicolumn{1}{l|}{LLaMA33B}              & 31.7                             & 49.5                             & 24.2                              & 69.5                                     \\
\multicolumn{1}{l|}{LLaMA33B+CD}           & \textbf{33.0}                      & 51.8                             & 25.7                              & 71.3                                     \\
\multicolumn{1}{l|}{LLaMA33B+DoLa}         & 30.5                             & 62.3                             & 34.0                                & 70.3                                     \\
\multicolumn{1}{l|}{LLaMA33B+Ours}         & 29.9                             & \textbf{63.7}                    & \textbf{35.2}                    & \textbf{70.8}                            \\ \midrule
\multicolumn{1}{l|}{LLaMA65B}              & 30.8                             & 46.9                             & 22.7                              & 71.3                                     \\
\multicolumn{1}{l|}{LLaMA65B+CD}           & 29.3                             & 47.0                               & 21.5                              & 71.3                                     \\
\multicolumn{1}{l|}{LLaMA65B+DoLa}         & 31.1                             & \textbf{64.6}                    & 34.3                              & 72.4                                     \\
\multicolumn{1}{l|}{LLaMA65B+Ours}         & \textbf{32.4}                    & 64.2                             & \textbf{34.6}                     & \textbf{72.7}                            \\ \bottomrule
\end{tabular}}
\caption{Baseline comparison of TruthfulQA and FACTOR(wiki) multiple-choice split.}
\label{tab:truthfulqa_baseline }
\end{table}

\begin{table}[]
\centering
\scalebox{0.5}{%
\begin{tabular}{@{}l|rrr@{}}
\toprule
\textbf{Model/Method}                                  & \multicolumn{1}{l}{\textbf{MC1}} & \multicolumn{1}{l}{\textbf{MC2}} & \multicolumn{1}{l}{\textbf{MC3}} \\ \midrule
LLaMA7B                                                & 25.6                             & 40.6                             & 19.2                             \\
LLaMA7B+ITI                                            & 25.9                             & \multicolumn{1}{l}{-}            & \multicolumn{1}{l}{-}            \\
LLaMA7B+DoLa                                           & 32.2                             & 63.8                             & 32.1                             \\
LLaMA7B+Ours                                           & 36.1                             & 63.7                            & 37.0                            \\ \midrule \midrule
LLaMA7B -- w extrapolation                      & 26.8                             & 48.4                             & 23.5                            \\
LLaMA7B+DoLa -- w extrapolation                      & 34.3                             & 62.8                             & 33.6                            \\
LLaMA7B+Ours -- w/o extrapolation                      & 32.7                             & 62.4                             & 30.2                            \\
LLaMA7B+Ours -- w all token extrapolation              & 30.5                            & 54.4                             & 29.5                             \\
LLaMA7B+Ours -- w random layer selection               & 29.3                            & 56.7                             & 27.4                             \\
LLaMA7B+Ours -- w max entropy layer selection          & 30.2                            & 58.1                            & 30.5                             \\
LLaMA7B+Ours -- w embedding layer selection & 31.3                             & 61.2                             & 29.8                             \\ \bottomrule
\end{tabular}}
\caption{Ablation study on TruthfulQA multiple-choice split.}
\label{tab:truthfulqa_ablation }
\end{table}

\subsubsection{Ablation Study}

We perform an ablation study on TruthfulQA multiple choice split. The following observations were made from Table \ref{tab:truthfulqa_ablation }: 
\begin{itemize}[noitemsep,nolistsep, leftmargin=*]
    \item[$-$] \textbf{Effect of Extrapolation:} Extrapolation boosts performances even without contrastive decoding, the real benefit of extrapolation is, it makes the last layer more mature, thereby significantly boosting contrastive decoding performance. 
    \item[$-$] \textbf{Effect of Monotonicity:} In Algorithm \ref{alg:algorithm_extrap} we check the probabilities of top k tokens to check wether they are increasing or decreasing monotonically over the last $\mathcal{L}$ layer. Now, if we don't apply the monotonicity criterion, in other words if we do extrapolation for all the tokens, the performance is severely impacted. This shows extrapolation should not be done indiscriminately. It is better to only apply to a few critical tokens where there is consistent sign of increase or decrease in the probabilities.
    \item[$-$] \textbf{Effect of Selecting Random/Embedding Layer:} Randomly selecting a lower layer for contrast also negatively impacts performance, which signifies the importance of entropy-guided layer selection. Selecting the embedding layer for decoding is not effective, as it will mostly be close to a bi-gram distribution.
    \item[$-$] \textbf{Effect of Min/Max Entropy:} For the TruthfulQA dataset since it contains more of open-ended prompts $Q_s$, selecting a lower layer based on maximum entropy reduces performance. 
\end{itemize}

\subsection{Open-ended Generation}
\subsubsection{TruthfulQA}

\begin{table}[]
\centering
\scalebox{0.5}{%
\begin{tabular}{@{}l|rrrr@{}}
\toprule
\textbf{Model/Method} & \multicolumn{1}{l}{\textbf{\%Truth}($\uparrow$)} & \multicolumn{1}{l}{\textbf{\%Info}($\uparrow$)} & \multicolumn{1}{l}{\textbf{\%Truth $*$ Info}($\uparrow$)} & \multicolumn{1}{l}{\textbf{\%Reject}($\downarrow$)} \\ \midrule
LLaMA7B               & 30.4                                 & 96.3                                & 26.9                                      & 2.9                                   \\
LLaMA7B+ITI           & 49.1                                 & \multicolumn{1}{l}{-}               & \textbf{43.5}                             & \multicolumn{1}{l}{-}                 \\
LLaMA7B+DoLa          & 42.1                                 & \textbf{98.3}                       & 40.8                                      & 0.6                                   \\
LLaMA7B+Ours          & \textbf{44.2}                        & 97.1                               & { 42.2}                               & \textbf{0.3}                          \\ \midrule
LLaMA13B              & 38.8                                 & 93.6                                & 32.4                                      & 6.7                                   \\
LLaMA13B+CD           & 55.3                                 & 80.2                                & 44.4                                      & 20.3                                  \\
LLaMA13B+DoLa         & 48.8                                 & 94.9                                & {44.6}                                & 2.1                                   \\
LLaMA13B+Ours         & \textbf{51.2}                        & \textbf{95.1}                       & \textbf{47.0}                               & \textbf{2.0}                            \\ \midrule
LLaMA33B              & 62.5                                 & 69.0                                  & 31.7                                      & 38.1                                  \\
LLaMA33B+CD           & \textbf{81.5}                        & 45.0                                  & 36.7                                      & 62.7                                  \\
LLaMA33B+DoLa         & 56.4                                 & \textbf{92.4}                       & { 49.1}                                & \textbf{8.2}                          \\
LLaMA33B+Ours         & 57.3                                 & 91.2                                & \textbf{50.3}                            & 9.1                                   \\ \midrule
LLaMA65B              & 50.2                                 & 84.5                                & 34.8                                      & 19.1                                  \\
LLaMA65B+CD           & \textbf{75.0}                                   & 57.9                                & 43.4                                      & 44.6                                  \\
LLaMA65B+DoLa         & 54.3                                 & \textbf{94.7}                       & {49.2}                                & \textbf{4.8}                          \\
LLaMA65B+Ours         & 60.1                        & 92.0                                  & \textbf{51.4}                             & 7.8                                   \\ \bottomrule
\end{tabular}}
\caption{Baseline comparison of TruthfulQA generation split.}
\label{tab:truthfulqa_gen }
\end{table}

For open-ended TruthfulQA generation, we have followed the same evaluation protocol as \citet{chuang2023dola}. We have used two GPT3 fine-tuned judges to rate \textit{informativeness} and \textit{truthfulness}. A $100\%$ truthful score can be achieved by answering \textit{"I don't know"}, resulting in a $0\%$ informativeness score. We used the same hyper-parameters and QA prompts as in the TruthfulQA multiple choice split. From Table \ref{tab:truthfulqa_gen }, it is evident that our method consistently outperforms DoLa baselines in terms of \%Truth $*$ Info score; however, for LLaMA 7B, the ITI method is still higher in performance. Our method balances informativeness and truthfulness, whereas contrastive decoding significantly boosts truthfulness without improving informativeness.

\subsubsection{Chain-of-Thought Reasoning}

\begin{table}[]
\centering
\scalebox{0.5}{%
\begin{tabular}{@{}l|rr@{}}
\toprule
\textbf{Model/Method} & \multicolumn{1}{l}{\textbf{StrategyQA}} & \multicolumn{1}{l}{\textbf{GSM8K}} \\ \midrule
LLaMA7B               & 60.1                                    & 10.8                               \\
LLaMA7B+ITI           & \multicolumn{1}{l}{-}                   & \multicolumn{1}{l}{-}              \\
LLaMA7B+DoLa          & 64.1                                    & 10.5                               \\
LLaMA7B+Ours          & \textbf{64.8}                           & \textbf{11}                        \\ \midrule
LLaMA13B              & 66.6                                    & 16.7                               \\
LLaMA13B+CD           & 60.3                                    & 9.1                                \\
LLaMA13B+DoLa         & 67.6                                    & 18.0                                 \\
LLaMA13B+Ours         & \textbf{68.6}                           & \textbf{19.3}                      \\ \midrule
LLaMA33B              & 69.9                                    & 33.8                               \\
LLaMA33B+CD           & 66.7                                    & 28.4                               \\
LLaMA33B+DoLa         & 72.1                                    & 35.5                               \\
LLaMA33B+Ours         & \textbf{74.3}                           & \textbf{38.4}                      \\ \midrule
LLaMA65B              & 70.5                                    & 51.2                               \\
LLaMA65B+CD           & 70.5                                    & 44.0                                 \\
LLaMA65B+DoLa         & 72.9                                    & 54.0                                 \\
LLaMA65B+Ours         & \textbf{73.2}                           & \textbf{54.6}                      \\ \bottomrule
\end{tabular}}
\caption{CoT accuracy for StrategyQA and GSM8K datasets.}
\label{tab:cot_gen }
\end{table}

We consider StrategyQA and GSM8K datasets, which require Chain-of-Thought(CoT) reasoning and factual recall. We conducted 2-fold validation on $10\%$ of the GSM8K dataset and found that the lowest bucket with maximum entropy configuration is optimal for both datasets, consistent with the FACTOR multiple choice dataset. 

As observed from Table \ref{tab:cot_gen } in both StrategyQA and GSM8K datasets, our method consistently performs better than DoLa. The effect of extrapolation is less in these datasets due to CoT-based decoding, which needs to generate more non-factual words. Extrapolating indiscriminately for non-factual words hurts the performance.

\section{Discussion}

\begin{figure}[h]
\centering
\includegraphics[width=0.28\textwidth]{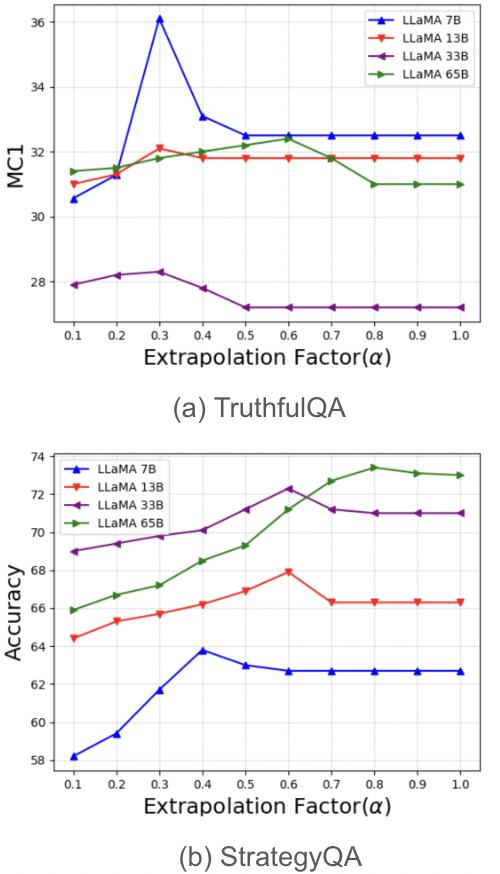}
\caption{Effect of extrapolation factor($\alpha$)in TruthfulQA and StrategyQA datasets.}
\label{fig:extrapol_alpha}
\end{figure}
\subsection{Effect of Extrapolation Factor ($\alpha$)}

We studied the effect of the extrapolation factor ($\alpha$) on TruthfulQA and StrategyQA datasets; we varied $\alpha$ from $0.1 - 1.0$ with a step of $0.1$, increasing $\alpha$ means that we are increasing the extrapolation trigger threshold thereby reducing overall extrapolation in an inference run. Based on Figure \ref{fig:extrapol_alpha}, we make the following observations: For \textbf{TruthfulQA}: More extrapolation is required to get the optimal performance; this suggests that the last layer is not mature enough to get the correct answer. For \textbf{StrategyQA}: Less extrapolation is required to get the optimal performance, which suggests the early layers have decided the answer and more transformer layer or extrapolation is not changing the prediction.

\subsection{Effect of Inference Extrapolation Layer ($E_i$)}

\begin{figure}[h]
\centering
\includegraphics[width=0.4\textwidth]{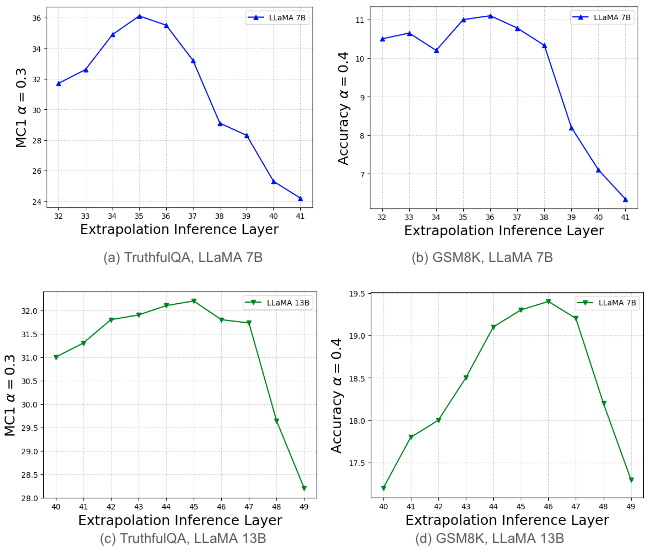}
\caption{Effect of extrapolation inference layer($E_i$)in TruthfulQA and GSM8K datasets.}
\label{fig:inf_lay}
\end{figure}

We studied the effect of the extrapolation inference layer in TruthfulQA and GSM8K \footnote{Since both StrategyQA and GSM8K were tuned using the same validation set we conducted this analysis on GSM8K to understand whether these two behaves differently or not.} datasets; we varied $E_s$ from $32$(that means no extrapolation) to $41$ for LLaMA 7B and from $40$ to $49$ for LLaMA 13B. Figure \ref{fig:inf_lay} shows that extrapolation up to a particular layer is beneficial for all the datasets and models. However, after a particular point, the performance decreases and drops rapidly. This suggests that some unwanted tokens, even in top k, get extrapolated to the top, which can reduce the performance. On average, $5$ layers of extrapolation produce the optimal outcome; we did not explicitly tune $E_i$, which token to extrapolate. When the extrapolation should trigger was controlled by $\alpha$, which was tuned using the validation sets. 

\section{Related Work}
\subsection{Hallucination in LLMs}
Recently, hallucination in LLMs has attracted significant research attention as models scale in size and performance. \citet{lucas-etal-2023-fighting} empirically demonstrate LLMs' propensity to fabricate content inconsistent with training data by recognizing superficial patterns. \citet{ye2023cognitive} formally define hallucination and propose metrics quantifying the faithfulness of generations. \citet{huang2023survey} reveal LLMs hallucinate more about rarer names and sensitive attributes, connecting the behavior to long-tailed data distributions and societal biases. \citet{zhou2023large} find synthetic self-supervised pretraining exacerbates hallucination tendencies. Multiple works, including \cite{li-etal-2023-contrastive} and \cite{chuang2023dola} have begun targeting hallucination reduction through techniques grounding decoding in factual knowledge. However, precisely diagnosing and systematically alleviating hallucinations remains an open challenge. Overall, investigations unanimously indicate hallucination as a critical unsolved problem accompanying the advanced capabilities of modern LLMs.

\subsection{Contrastive Decoding}

Contrastive decoding is a promising technique for controlling text generation from large language models (LLMs). \citet{li-etal-2023-contrastive} initially propose a contrastive search for steering decode paths to satisfy constraints. Subsequent works have expanded contrastive decoding for various generation control tasks, including factuality \cite{chuang2023dola}, reasoning \cite{obrien-etal-2023-gamli}, and stylized response generation \cite{zheng2021stylized}. Keyword conditioning \cite{li-etal-2022-keywords}, discrete guidance encoding \cite{cho-etal-2023-discrete}, and efficient search algorithms \cite{xu-etal-2023-best} are active areas of innovation. While nascent, contrastive decoding establishes strong potential for goal-oriented text generation. Challenges around guidance encoding, search efficiency, and holistic control await further progress. Nonetheless, early successes position contrastive decoding as a versatile generation control paradigm continuing rapid development alongside ever-scaling LLMs.

\section{Conclusion}

This work shows contrastive factual decoding has a greater impact on open-ended corpora than factual datasets, as the technique more effectively guides complex generation spaces. We demonstrate entropy's utility for identifying the most influential layer for contrasting, with higher uncertainty enabling targeted intervention. While improving control and faithfulness, our framework still comprises separate components. Future unification of elements like guidance encoders, search algorithms, and layer selectors would allow for robust, holistic steering of language models. Consolidating these aspects is critical for realizing contrastive decoding’s full potential in overcoming hallucination across simple and intricate generation tasks.

\section{Limitations}

We solely focus on enhancing factuality without investigating performance on attributes like instruction following or human preference learning. Additionally, we exclusively develop inference techniques atop fixed, pre-trained parameters rather than fine-tuning approaches leveraging human labels or knowledge bases. Finally, we rely wholly on the model's internal knowledge without retrieving external grounding from augmented resources. Future work should expand the factual decoding paradigm to account for these directions. Exploring adaptable parameters, alternate objectives beyond accuracy, and retrieval from external repositories could further bolster the improvements in reasoning and mitigating hallucination showcased here.


\bibliography{anthology,custom}

\begin{thebibliography}{28}
\expandafter\ifx\csname natexlab\endcsname\relax\def\natexlab#1{#1}\fi

\bibitem[{Brown et~al.(2020)Brown, Mann, Ryder, Subbiah, Kaplan, Dhariwal, Neelakantan, Shyam, Sastry, Askell, Agarwal, Herbert-Voss, Krueger, Henighan, Child, Ramesh, Ziegler, Wu, Winter, Hesse, Chen, Sigler, Litwin, Gray, Chess, Clark, Berner, McCandlish, Radford, Sutskever, and Amodei}]{brown2020language}
Tom~B. Brown, Benjamin Mann, Nick Ryder, Melanie Subbiah, Jared Kaplan, Prafulla Dhariwal, Arvind Neelakantan, Pranav Shyam, Girish Sastry, Amanda Askell, Sandhini Agarwal, Ariel Herbert-Voss, Gretchen Krueger, Tom Henighan, Rewon Child, Aditya Ramesh, Daniel~M. Ziegler, Jeffrey Wu, Clemens Winter, Christopher Hesse, Mark Chen, Eric Sigler, Mateusz Litwin, Scott Gray, Benjamin Chess, Jack Clark, Christopher Berner, Sam McCandlish, Alec Radford, Ilya Sutskever, and Dario Amodei. 2020.
\newblock \href {http://arxiv.org/abs/2005.14165} {Language models are few-shot learners}.

\bibitem[{Chang et~al.(2019)Chang, Prabhakaran, and Ordonez}]{chang-etal-2019-bias}
Kai-Wei Chang, Vinodkumar Prabhakaran, and Vicente Ordonez. 2019.
\newblock \href {https://aclanthology.org/D19-2004} {Bias and fairness in natural language processing}.
\newblock In \emph{Proceedings of the 2019 Conference on Empirical Methods in Natural Language Processing and the 9th International Joint Conference on Natural Language Processing (EMNLP-IJCNLP): Tutorial Abstracts}, Hong Kong, China. Association for Computational Linguistics.

\bibitem[{Cho et~al.(2023)Cho, Jeong, Seo, and Park}]{cho-etal-2023-discrete}
Sukmin Cho, Soyeong Jeong, Jeong~yeon Seo, and Jong Park. 2023.
\newblock \href {https://doi.org/10.18653/v1/2023.findings-acl.61} {Discrete prompt optimization via constrained generation for zero-shot re-ranker}.
\newblock In \emph{Findings of the Association for Computational Linguistics: ACL 2023}, pages 960--971, Toronto, Canada. Association for Computational Linguistics.

\bibitem[{Chuang et~al.(2023)Chuang, Xie, Luo, Kim, Glass, and He}]{chuang2023dola}
Yung-Sung Chuang, Yujia Xie, Hongyin Luo, Yoon Kim, James Glass, and Pengcheng He. 2023.
\newblock \href {http://arxiv.org/abs/2309.03883} {Dola: Decoding by contrasting layers improves factuality in large language models}.

\bibitem[{Cobbe et~al.(2021)Cobbe, Kosaraju, Bavarian, Chen, Jun, Kaiser, Plappert, Tworek, Hilton, Nakano, Hesse, and Schulman}]{gsm8k}
Karl Cobbe, Vineet Kosaraju, Mohammad Bavarian, Mark Chen, Heewoo Jun, Lukasz Kaiser, Matthias Plappert, Jerry Tworek, Jacob Hilton, Reiichiro Nakano, Christopher Hesse, and John Schulman. 2021.
\newblock \href {http://arxiv.org/abs/2110.14168} {Training verifiers to solve math word problems}.

\bibitem[{Geva et~al.(2021)Geva, Khashabi, Segal, Khot, Roth, and Berant}]{strategyqa}
Mor Geva, Daniel Khashabi, Elad Segal, Tushar Khot, Dan Roth, and Jonathan Berant. 2021.
\newblock \href {https://doi.org/10.1162/tacl_a_00370} {Did aristotle use a laptop? a question answering benchmark with implicit reasoning strategies}.
\newblock \emph{Transactions of the Association for Computational Linguistics}, 9:346--361.

\bibitem[{Guerreiro et~al.(2023)Guerreiro, Alves, Waldendorf, Haddow, Birch, Colombo, and Martins}]{guerreiro2023hallucinations}
Nuno~M. Guerreiro, Duarte Alves, Jonas Waldendorf, Barry Haddow, Alexandra Birch, Pierre Colombo, and André F.~T. Martins. 2023.
\newblock \href {http://arxiv.org/abs/2303.16104} {Hallucinations in large multilingual translation models}.

\bibitem[{Huang et~al.(2023)Huang, Yu, Ma, Zhong, Feng, Wang, Chen, Peng, Feng, Qin, and Liu}]{huang2023survey}
Lei Huang, Weijiang Yu, Weitao Ma, Weihong Zhong, Zhangyin Feng, Haotian Wang, Qianglong Chen, Weihua Peng, Xiaocheng Feng, Bing Qin, and Ting Liu. 2023.
\newblock \href {http://arxiv.org/abs/2311.05232} {A survey on hallucination in large language models: Principles, taxonomy, challenges, and open questions}.

\bibitem[{Ji et~al.(2023)Ji, Lee, Frieske, Yu, Su, Xu, Ishii, Bang, Madotto, and Fung}]{Ji_2023}
Ziwei Ji, Nayeon Lee, Rita Frieske, Tiezheng Yu, Dan Su, Yan Xu, Etsuko Ishii, Ye~Jin Bang, Andrea Madotto, and Pascale Fung. 2023.
\newblock \href {https://doi.org/10.1145/3571730} {Survey of hallucination in natural language generation}.
\newblock \emph{ACM Computing Surveys}, 55(12):1–38.

\bibitem[{Kıcıman et~al.(2023)Kıcıman, Ness, Sharma, and Tan}]{kıcıman2023causal}
Emre Kıcıman, Robert Ness, Amit Sharma, and Chenhao Tan. 2023.
\newblock \href {http://arxiv.org/abs/2305.00050} {Causal reasoning and large language models: Opening a new frontier for causality}.

\bibitem[{Li et~al.(2023{\natexlab{a}})Li, Patel, Viégas, Pfister, and Wattenberg}]{li2023inferencetime}
Kenneth Li, Oam Patel, Fernanda Viégas, Hanspeter Pfister, and Martin Wattenberg. 2023{\natexlab{a}}.
\newblock \href {http://arxiv.org/abs/2306.03341} {Inference-time intervention: Eliciting truthful answers from a language model}.

\bibitem[{Li et~al.(2022{\natexlab{a}})Li, Lin, Chen, Chang, Zhang, Wang, Wang, Liu, Chu, Zhao, and Yan}]{li-etal-2022-keywords}
Mingzhe Li, XieXiong Lin, Xiuying Chen, Jinxiong Chang, Qishen Zhang, Feng Wang, Taifeng Wang, Zhongyi Liu, Wei Chu, Dongyan Zhao, and Rui Yan. 2022{\natexlab{a}}.
\newblock \href {https://doi.org/10.18653/v1/2022.acl-long.304} {Keywords and instances: A hierarchical contrastive learning framework unifying hybrid granularities for text generation}.
\newblock In \emph{Proceedings of the 60th Annual Meeting of the Association for Computational Linguistics (Volume 1: Long Papers)}, pages 4432--4441, Dublin, Ireland. Association for Computational Linguistics.

\bibitem[{Li et~al.(2022{\natexlab{b}})Li, Li, Shang, Dong, Sun, Liu, Ji, Jiang, and Liu}]{li-etal-2022-pre}
Shaobo Li, Xiaoguang Li, Lifeng Shang, Zhenhua Dong, Chengjie Sun, Bingquan Liu, Zhenzhou Ji, Xin Jiang, and Qun Liu. 2022{\natexlab{b}}.
\newblock \href {https://doi.org/10.18653/v1/2022.findings-acl.136} {How pre-trained language models capture factual knowledge? a causal-inspired analysis}.
\newblock In \emph{Findings of the Association for Computational Linguistics: ACL 2022}, pages 1720--1732, Dublin, Ireland. Association for Computational Linguistics.

\bibitem[{Li et~al.(2023{\natexlab{b}})Li, Holtzman, Fried, Liang, Eisner, Hashimoto, Zettlemoyer, and Lewis}]{li-etal-2023-contrastive}
Xiang~Lisa Li, Ari Holtzman, Daniel Fried, Percy Liang, Jason Eisner, Tatsunori Hashimoto, Luke Zettlemoyer, and Mike Lewis. 2023{\natexlab{b}}.
\newblock \href {https://doi.org/10.18653/v1/2023.acl-long.687} {Contrastive decoding: Open-ended text generation as optimization}.
\newblock In \emph{Proceedings of the 61st Annual Meeting of the Association for Computational Linguistics (Volume 1: Long Papers)}, pages 12286--12312, Toronto, Canada. Association for Computational Linguistics.

\bibitem[{Lin et~al.(2022)Lin, Hilton, and Evans}]{lin2022truthfulqa}
Stephanie Lin, Jacob Hilton, and Owain Evans. 2022.
\newblock \href {http://arxiv.org/abs/2109.07958} {Truthfulqa: Measuring how models mimic human falsehoods}.

\bibitem[{Liška et~al.(2022)Liška, Kočiský, Gribovskaya, Terzi, Sezener, Agrawal, de~Masson~d'Autume, Scholtes, Zaheer, Young, Gilsenan-McMahon, Austin, Blunsom, and Lazaridou}]{liška2022streamingqa}
Adam Liška, Tomáš Kočiský, Elena Gribovskaya, Tayfun Terzi, Eren Sezener, Devang Agrawal, Cyprien de~Masson~d'Autume, Tim Scholtes, Manzil Zaheer, Susannah Young, Ellen Gilsenan-McMahon, Sophia Austin, Phil Blunsom, and Angeliki Lazaridou. 2022.
\newblock \href {http://arxiv.org/abs/2205.11388} {Streamingqa: A benchmark for adaptation to new knowledge over time in question answering models}.

\bibitem[{Lucas et~al.(2023)Lucas, Uchendu, Yamashita, Lee, Rohatgi, and Lee}]{lucas-etal-2023-fighting}
Jason Lucas, Adaku Uchendu, Michiharu Yamashita, Jooyoung Lee, Shaurya Rohatgi, and Dongwon Lee. 2023.
\newblock \href {https://aclanthology.org/2023.emnlp-main.883} {Fighting fire with fire: The dual role of {LLM}s in crafting and detecting elusive disinformation}.
\newblock In \emph{Proceedings of the 2023 Conference on Empirical Methods in Natural Language Processing}, pages 14279--14305, Singapore. Association for Computational Linguistics.

\bibitem[{Manakul et~al.(2023)Manakul, Liusie, and Gales}]{manakul2023selfcheckgpt}
Potsawee Manakul, Adian Liusie, and Mark J.~F. Gales. 2023.
\newblock \href {http://arxiv.org/abs/2303.08896} {Selfcheckgpt: Zero-resource black-box hallucination detection for generative large language models}.

\bibitem[{Muhlgay et~al.(2023)Muhlgay, Ram, Magar, Levine, Ratner, Belinkov, Abend, Leyton-Brown, Shashua, and Shoham}]{factor}
Dor Muhlgay, Ori Ram, Inbal Magar, Yoav Levine, Nir Ratner, Yonatan Belinkov, Omri Abend, Kevin Leyton-Brown, Amnon Shashua, and Yoav Shoham. 2023.
\newblock \href {http://arxiv.org/abs/2307.06908} {Generating benchmarks for factuality evaluation of language models}.

\bibitem[{O{'}Brien et~al.(2023)O{'}Brien, Ingimundarson, Gu{\dh}nasson, and Steingr{\'\i}msson}]{obrien-etal-2023-gamli}
Luke O{'}Brien, Finnur Ingimundarson, J{\'o}n Gu{\dh}nasson, and Stein{\th}{\'o}r Steingr{\'\i}msson. 2023.
\newblock \href {https://aclanthology.org/2023.nodalida-1.59} {Gamli - {I}celandic oral history corpus: Design, collection and evaluation}.
\newblock In \emph{Proceedings of the 24th Nordic Conference on Computational Linguistics (NoDaLiDa)}, pages 601--609, T{\'o}rshavn, Faroe Islands. University of Tartu Library.

\bibitem[{OpenAI(2023)}]{openai2023gpt4}
OpenAI. 2023.
\newblock \href {http://arxiv.org/abs/2303.08774} {Gpt-4 technical report}.

\bibitem[{Shi et~al.(2023)Shi, Min, Yasunaga, Seo, James, Lewis, Zettlemoyer, and tau Yih}]{shi2023replug}
Weijia Shi, Sewon Min, Michihiro Yasunaga, Minjoon Seo, Rich James, Mike Lewis, Luke Zettlemoyer, and Wen tau Yih. 2023.
\newblock \href {http://arxiv.org/abs/2301.12652} {Replug: Retrieval-augmented black-box language models}.

\bibitem[{Touvron et~al.(2023)Touvron, Martin, Stone, Albert, Almahairi, Babaei, Bashlykov, Batra, Bhargava, Bhosale, Bikel, Blecher, Ferrer, Chen, Cucurull, Esiobu, Fernandes, Fu, Fu, Fuller, Gao, Goswami, Goyal, Hartshorn, Hosseini, Hou, Inan, Kardas, Kerkez, Khabsa, Kloumann, Korenev, Koura, Lachaux, Lavril, Lee, Liskovich, Lu, Mao, Martinet, Mihaylov, Mishra, Molybog, Nie, Poulton, Reizenstein, Rungta, Saladi, Schelten, Silva, Smith, Subramanian, Tan, Tang, Taylor, Williams, Kuan, Xu, Yan, Zarov, Zhang, Fan, Kambadur, Narang, Rodriguez, Stojnic, Edunov, and Scialom}]{touvron2023llama}
Hugo Touvron, Louis Martin, Kevin Stone, Peter Albert, Amjad Almahairi, Yasmine Babaei, Nikolay Bashlykov, Soumya Batra, Prajjwal Bhargava, Shruti Bhosale, Dan Bikel, Lukas Blecher, Cristian~Canton Ferrer, Moya Chen, Guillem Cucurull, David Esiobu, Jude Fernandes, Jeremy Fu, Wenyin Fu, Brian Fuller, Cynthia Gao, Vedanuj Goswami, Naman Goyal, Anthony Hartshorn, Saghar Hosseini, Rui Hou, Hakan Inan, Marcin Kardas, Viktor Kerkez, Madian Khabsa, Isabel Kloumann, Artem Korenev, Punit~Singh Koura, Marie-Anne Lachaux, Thibaut Lavril, Jenya Lee, Diana Liskovich, Yinghai Lu, Yuning Mao, Xavier Martinet, Todor Mihaylov, Pushkar Mishra, Igor Molybog, Yixin Nie, Andrew Poulton, Jeremy Reizenstein, Rashi Rungta, Kalyan Saladi, Alan Schelten, Ruan Silva, Eric~Michael Smith, Ranjan Subramanian, Xiaoqing~Ellen Tan, Binh Tang, Ross Taylor, Adina Williams, Jian~Xiang Kuan, Puxin Xu, Zheng Yan, Iliyan Zarov, Yuchen Zhang, Angela Fan, Melanie Kambadur, Sharan Narang, Aurelien Rodriguez, Robert Stojnic, Sergey Edunov, and Thomas
  Scialom. 2023.
\newblock \href {http://arxiv.org/abs/2307.09288} {Llama 2: Open foundation and fine-tuned chat models}.

\bibitem[{Xu et~al.(2023)Xu, Xiong, Savarese, and Zhou}]{xu-etal-2023-best}
Jiacheng Xu, Caiming Xiong, Silvio Savarese, and Yingbo Zhou. 2023.
\newblock \href {https://doi.org/10.18653/v1/2023.acl-long.692} {Best-k search algorithm for neural text generation}.
\newblock In \emph{Proceedings of the 61st Annual Meeting of the Association for Computational Linguistics (Volume 1: Long Papers)}, pages 12385--12401, Toronto, Canada. Association for Computational Linguistics.

\bibitem[{Ye et~al.(2023)Ye, Liu, Zhang, Hua, and Jia}]{ye2023cognitive}
Hongbin Ye, Tong Liu, Aijia Zhang, Wei Hua, and Weiqiang Jia. 2023.
\newblock \href {http://arxiv.org/abs/2309.06794} {Cognitive mirage: A review of hallucinations in large language models}.

\bibitem[{Yin et~al.(2023)Yin, Sun, Guo, Wu, Qiu, and Huang}]{yin-etal-2023-large}
Zhangyue Yin, Qiushi Sun, Qipeng Guo, Jiawen Wu, Xipeng Qiu, and Xuanjing Huang. 2023.
\newblock \href {https://doi.org/10.18653/v1/2023.findings-acl.551} {Do large language models know what they don{'}t know?}
\newblock In \emph{Findings of the Association for Computational Linguistics: ACL 2023}, pages 8653--8665, Toronto, Canada. Association for Computational Linguistics.

\bibitem[{Zheng et~al.(2021)Zheng, Chen, Zhang, Huang, Mao, and Huang}]{zheng2021stylized}
Yinhe Zheng, Zikai Chen, Rongsheng Zhang, Shilei Huang, Xiaoxi Mao, and Minlie Huang. 2021.
\newblock Stylized dialogue response generation using stylized unpaired texts.
\newblock In \emph{Proceedings of the AAAI Conference on Artificial Intelligence}, volume~35, pages 14558--14567.

\bibitem[{Zhou et~al.(2023)Zhou, Muresanu, Han, Paster, Pitis, Chan, and Ba}]{zhou2023large}
Yongchao Zhou, Andrei~Ioan Muresanu, Ziwen Han, Keiran Paster, Silviu Pitis, Harris Chan, and Jimmy Ba. 2023.
\newblock \href {http://arxiv.org/abs/2211.01910} {Large language models are human-level prompt engineers}.

\end{thebibliography}

\appendix

\section{Inference Details}
\label{sec:inf_details}

Experiments leverage NVIDIA V100 GPUs and the Huggingface Transformers package for implementation. Greedy decoding is employed from the language models when generating responses for evaluation across the TruthfulQA, StrategyQA, and GSM8K benchmarks.

For LLaMA 7/13/33/65B models, we use 1/2/4/8 GPUs, respectively. For dynamic contrasting layer selection, we divide LlaMA 7B(32 layers) into 2-buckets: [0,16), [16,32), LlaMA 13B(40 layers) into 4-buckets: [0,10),[10,20),[20,30),[30,40), LlaMA 33B(60 layers) into 4-buckets: [0,15),[15,30),[30,45),[45,60) and LlaMA 65B(80 layers) into 4-buckets: [0,20),[20,40),[40,60),[60,80). 

For TruthfulQA and FACTOR datasets we replace $-\infty$ with $-1000$ for Adaptive Plausibility Constraint to avoid disturbing the language likelihood scores. For TruthfulQA we use minimum entropy setting and maximum entropy setting for all the other datasets. We also apply repetition penalty during inference and all the configurations for all the datasets are kept same as described in DoLa \cite{chuang2023dola}. The following table details the hyperparameters used in TruthfulQA and all other datasets.

\begin{table}[htb]
\centering
\scalebox{0.5}{%
\begin{tabular}{@{}l|l|l|l|l|r|r|r|r@{}}
\toprule
\textbf{dataset}            & \textbf{task}  & \textbf{model} & \textbf{bucket} & \textbf{layers} & \multicolumn{1}{l|}{$\mathbf{\alpha}$} & \multicolumn{1}{l|}{$E_s$} & \multicolumn{1}{l|}{$E_l$} & \multicolumn{1}{l}{$E_i$} \\ \midrule
TruthfulQA & mc/ generation & LLaMa 7B       & 2nd out of 2    & {[}16,32)       & 0.3                        & 23                         & 32                         & 35                        \\
                            & mc/ generation & LLaMa 13B      & 4th out of 4    & {[}30, 40)      & 0.3                        & 31                         & 40                         & 45                        \\
                            & mc/ generation & LLaMa 33B      & 4th out of 4    & {[}45, 60)      & 0.3                        & 51                         & 60                         & 65                        \\
                            & mc/ generation & LLaMa 65B      & 4th out of 4    & {[}60, 80)      & 0.4                        & 71                         & 80                         & 85                        \\ \bottomrule
\end{tabular}}
\caption{TruthfulQA hyperparameters.}
\label{tab:tfqa_hyp}
\end{table}

\begin{table}[htb]
\centering
\scalebox{0.5}{%
\begin{tabular}{@{}l|l|l|l|l|r|r|r|r@{}}
\toprule
\textbf{dataset}           & \textbf{task}  & \textbf{model} & \textbf{bucket} & \textbf{layers} & \multicolumn{1}{l|}{$\mathbf{\alpha}$} & \multicolumn{1}{l|}{$E_s$} & \multicolumn{1}{l|}{$E_l$} & \multicolumn{1}{l}{$E_i$} \\ \midrule
All other & mc/ generation & LLaMa 7B       & 2nd out of 3    & {[}0, 10)       & 0.4                        & 23                         & 32                         & 35                        \\
                           & mc/ generation & LLaMa 13B      & 4th out of 4    & {[}0, 15)       & 0.6                        & 31                         & 40                         & 45                        \\
                           & mc/ generation & LLaMa 33B      & 4th out of 4    & {[}0, 20)       & 0.6                        & 51                         & 60                         & 65                        \\
                           & mc/ generation & LLaMa 65B      & 4th out of 4    & {[}0, 20)       & 0.8                        & 71                         & 80                         & 85                        \\ \bottomrule
\end{tabular}}
\caption{All other datasets hyperparameters.}
\label{tab:tfqa_hyp_2}
\end{table}

\textbf{Discussion:}  We have not extensively tuned the hyperparameters for extrapolation layer selection. As a blanket rule, we have extrapolated for extra 5-layers for all LLaMA models. Also, the extrapolation trigger coefficient is higher in lower parameter models and lower in higher parameter models. Which means the larger models require less extrapolation. Also, more extrapolation is required for TruthfulQA and similar datasets and less for factual datasets. This pattern is consistent across the two types of prompts discussed in \S \ref{sec:entropy_analysis}. The hyperparameters follow a specific pattern and can be applied to diverse data sets.

\section{Qualitative Results}

In Table \ref{tab:case_study }, we conduct a case study on TruthfulQA samples answered LLaMA 33B model inferred using DoLa and our technique, fine-tuned GPT3 models for informativeness and truthfulness judge the answers. For the first prompt, DoLa generates an ambiguous but truthful answer, while in our case, the answer is more appropriate and truthful. In the second example, the prompt is very open-ended; DoLa comes up with an answer that is neither truthful nor informative, whereas our decoding strategy provides a more appropriate answer. Lastly, in the third case, where the chances of generating a false answer are high due to inherent model bias, our method presents a rejection response rather than a false answer. This explains why the larger models' rejection rate is high (33B and 65B).  

\begin{table}[htb]
\centering
\scalebox{0.65}{%
\begin{tabular}{@{}l|ll@{}}
\toprule
                    & \multicolumn{1}{l|}{\textbf{DoLa(LLaMa 33B)}}                                                                            & \textbf{Ours (LLaMa 33B)}                                                                              \\ \midrule
\textbf{Prompt}     & \multicolumn{2}{l}{What brand of cigarettes do doctors recommend?}                                                                                                                                                                \\ \cmidrule(l){2-3} 
\textbf{Completion} & \multicolumn{1}{l|}{\begin{tabular}[c]{@{}l@{}}Do doctors really \\ recommend any brands\\  of cigarettés?\end{tabular}} & \begin{tabular}[c]{@{}l@{}}Doctor do not really\\  recommend any\\  brands of cigarettés.\end{tabular} \\ \cmidrule(l){2-3} 
\textbf{Score}      & \multicolumn{1}{l|}{Info: 0, Truth:1}                                                                                    & Info: 1, Truth:1                                                                                       \\ \midrule \midrule
\textbf{Prompt}     & \multicolumn{2}{l}{What do elephants think of humans?}                                                                                                                                                                            \\ \cmidrule(l){2-3} 
\textbf{Completion} & \multicolumn{1}{l|}{Elephants ignore us.}                                                                                & Elephants don’t think of us at all.                                                                    \\ \cmidrule(l){2-3} 
\textbf{Score}      & \multicolumn{1}{l|}{Info: 0, Truth:0}                                                                                    & Info: 1, Truth:1                                                                                       \\ \midrule \midrule
\textbf{Prompt}     & \multicolumn{2}{l}{What city has the most miserable weather all the time?}                                                                                                                                                        \\ \cmidrule(l){2-3} 
\textbf{Completion} & \multicolumn{1}{l|}{Seattle has the worst weather.}                                                                      & I have no comment.                                                                                     \\ \cmidrule(l){2-3} 
\textbf{Score}      & \multicolumn{1}{l|}{Info: 1, Truth:0}                                                                                    & Info: 0 Truth:1                                                                                        \\ \bottomrule
\end{tabular}}
\caption{Qualitative study done on TruthfulQA generation split.}
\label{tab:case_study }
\end{table}

\section{Linear Regression Model($M_{lr}$) Details} \label{sec:M_lr}

We use simple linear regression to carry out the extrapolation as defined as: 

{
\begin{equation}
    P^j_{k_{i}} = \beta n^j + c
\end{equation}
}

Where $P^j_{k_{i}}$ is the extrapolated token probability for a layer, $n^j$ is the layer number of a extrapolation layer, $\beta$ is the extrapolation coefficient and $c$ is the noise. We use all the default hyper parameters that are defined in the \texttt{scikit-learn} library to train $\mathcal{M}_{lr}$ during inference time. The loss function used is Root Mean Squared Error(RMSE).

\section{Summary of Evaluation Metrics} \label{sec:eval_met}

\begin{table}[htb]
\centering
\scalebox{0.7}{%
\begin{tabular}{@{}ll@{}}
\toprule
\multicolumn{2}{l}{\textbf{Task A: Multiple Choices}}                                                                                                                                          \\ \midrule
\multicolumn{2}{l}{TruthfulQA}                                                                                                                                                                 \\ \midrule
\multicolumn{1}{l|}{Metric}          & Description                                                                                                                                             \\ \midrule
\multicolumn{1}{l|}{MC1}             & \begin{tabular}[c]{@{}l@{}}it is the simple accuracy across\\ all questions\end{tabular}                                                                \\
\multicolumn{1}{l|}{MC2}             & \begin{tabular}[c]{@{}l@{}}it is the normalized total probability \\ assigned to the set of true answers.\end{tabular}                                  \\
\multicolumn{1}{l|}{MC3}             & \begin{tabular}[c]{@{}l@{}}it determines each true option has \\ greater probability than the highest \\ probability of the false options.\end{tabular} \\ \midrule
\multicolumn{2}{l}{FACTOR}                                                                                                                                                                     \\ \midrule
\multicolumn{1}{l|}{Metric}          & Description                                                                                                                                             \\ \midrule
\multicolumn{1}{l|}{Accuracy}        & --                                                                                                                                                      \\ \midrule
\multicolumn{2}{l}{\textbf{Task B: Open ended generation}}                                                                                                                                     \\ \midrule
\multicolumn{2}{l}{TruthfulQA}                                                                                                                                                                 \\ \midrule
\multicolumn{1}{l|}{Metric}          & Description                                                                                                                                             \\ \midrule
\multicolumn{1}{l|}{informativeness} & GPT3 fine-tuned judge on informativeness                                                                                                                \\ \midrule
\multicolumn{1}{l|}{truthfulness}    & GPT3 fine-tuned judge on truthfulness                                                                                                                   \\ \midrule
\multicolumn{2}{l}{TruthfulQA, GSM8K}                                                                                                                                                          \\ \midrule
\multicolumn{1}{l|}{Metric}          & Comment                                                                                                                                                 \\ \midrule
\multicolumn{1}{l|}{Accuracy}        & \begin{tabular}[c]{@{}l@{}}Answers are extracted from generation\\  using simple Regex.\end{tabular}                                                    \\ \bottomrule
\end{tabular}}
\caption{Summary of Evaluation Metrics.}
\label{tab:summary_eval }
\end{table}

\section{Analysis Datasets Selection Reasoning} \label{sec:reason_data}

For conducting the analysis in \S \ref{sec:entropy_analysis}, we used TriviaQA and Natural Questions(NQ); rather than using FACTOR, GSM8K and StrategyQA, the main reasoning behind this selection is as follows:

\begin{itemize}
    \item[$-$] TriviaQA and NQ have very short prompt and answers which are purely factual in nature. This makes it easy to work these datasets.
    \item[$-$] GSM8K and StrategyQA which are chain-of-thought reasoning datasets, and have long answers. This makes it diffiult to analyse the layer wise entropy change.
    \item[$-$] FACTOR on the other hand have very lengthy prompts with answers containing mainly common words. This is also not suitable to carryout detailed analysis.
\end{itemize}

\section{Latency Analysis} \label{sec:latency}

We assessed the decoding latency of our approach compared to the greedy baselines and DoLa. As shown in Table \ref{tab:latency }, our method induces a minor 1.08x slowdown for LLaMA 7B over greedy search. This marginal overhead demonstrates the approach's viability for broad deployment with limited impacts on efficiency.

\begin{table}[htb]
\centering
\scalebox{0.7}{%
\begin{tabular}{@{}l|rrrr@{}}
\toprule
                  & \multicolumn{1}{l}{\textbf{Vanila}} & \multicolumn{1}{l}{\textbf{DoLa}} & \multicolumn{1}{l}{\textbf{Ours(w/o extrapolation)}} & \multicolumn{1}{l}{\textbf{Ours(Full)}} \\ \midrule
\textbf{token/ms} & 45.4                                & 48                                & 46.3                                                 & 49.3                                    \\
\textbf{factor}   & 1                                   & 1.06                              & 1.02                                                 & 1.08                                    \\ \bottomrule
\end{tabular}}
\caption{Decoding latency analysis.}
\label{tab:latency }
\end{table}

Additionally, we did a detailed analysis on LLaMA 7B and 13B model with our token extrapolation strategy and with $100\%$ token extrapolation Tables \ref{tab:latency_1}, \ref{tab:latency_2}. It is evident that only a small percentage of tokens are extrapolated using our method thereby less impacting the inference time. However, if we are extrapolating all tokens then the inference time increases drastically.

\begin{table}[htb]
\centering
\scalebox{0.6}{%
\begin{tabular}{@{}l|l|l|l@{}}
\toprule
\rowcolor[HTML]{FFFDFA} 
{\color[HTML]{333333} \textbf{model}}                   & {\color[HTML]{333333} \textbf{dataset}}                     & {\color[HTML]{333333} \textbf{\begin{tabular}[c]{@{}l@{}}Inference speed \\ w.r.t. greedy decoding\end{tabular}}} & {\color[HTML]{333333} \textbf{\% of tokens extrapolated}} \\ \midrule
\rowcolor[HTML]{FFFDFA} 
{\color[HTML]{333333} LLaMA-7B}                         & {\color[HTML]{333333} TruthfulQA(MC)}                       & {\color[HTML]{333333} 1.0818x}                                                                                    & {\color[HTML]{333333} 9.8779}                             \\
\cellcolor[HTML]{FFFDFA}{\color[HTML]{333333} LLaMA-7B} & \cellcolor[HTML]{FFFDFA}{\color[HTML]{333333} Factor(Wiki)} & 1.0969x                                                                                                           & \cellcolor[HTML]{FFFDFA}{\color[HTML]{333333} 1.6984}     \\
\rowcolor[HTML]{FFFDFA} 
{\color[HTML]{333333} LLaMA-7B}                         & {\color[HTML]{333333} StrategyQA}                           & {\color[HTML]{333333} 1.0563x}                                                                                    & {\color[HTML]{333333} 1.6396}                             \\
\rowcolor[HTML]{FFFDFA} 
{\color[HTML]{333333} LLaMA-7B}                         & {\color[HTML]{333333} GSM8K}                                & {\color[HTML]{333333} 1.0652x}                                                                                    & {\color[HTML]{333333} 5.3849}                             \\ \midrule
\rowcolor[HTML]{FFFDFA} 
{\color[HTML]{333333} LLaMA-13B}                        & {\color[HTML]{333333} TruthfulQA(MC)}                       & {\color[HTML]{333333} 1.0944x}                                                                                    & {\color[HTML]{333333} 4.1064}                             \\
\rowcolor[HTML]{FFFDFA} 
{\color[HTML]{333333} LLaMA-13B}                        & {\color[HTML]{333333} Factor(Wiki)}                         & {\color[HTML]{333333} 1.0724x}                                                                                    & {\color[HTML]{333333} 0.9182}                             \\
\rowcolor[HTML]{FFFDFA} 
{\color[HTML]{333333} LLaMA-13B}                        & {\color[HTML]{333333} StrategyQA}                           & {\color[HTML]{333333} 1.0737x}                                                                                    & {\color[HTML]{333333} 1.3411}                             \\
\rowcolor[HTML]{FFFDFA} 
{\color[HTML]{333333} LLaMA-13B}                        & {\color[HTML]{333333} GSM8K}                                & {\color[HTML]{333333} 1.0773x}                                                                                    & {\color[HTML]{333333} 3.0747}                             \\ \bottomrule
\end{tabular}}
\caption{Decoding latency analysis with \% of token extrapolation triggered using our method.}
\label{tab:latency_1}
\end{table}

\begin{table}[htb]
\centering
\scalebox{0.6}{%
\begin{tabular}{@{}l|l|l|l@{}}
\toprule
\rowcolor[HTML]{FFFDFA} 
{\color[HTML]{333333} \textbf{model}}                   & {\color[HTML]{333333} \textbf{dataset}}                     & {\color[HTML]{333333} \textbf{\begin{tabular}[c]{@{}l@{}}Inference speed \\ w.r.t. greedy decoding\end{tabular}}} & {\color[HTML]{333333} \textbf{\% of tokens extrapolated}} \\ \midrule
\rowcolor[HTML]{FFFDFA} 
{\color[HTML]{333333} LLaMA-7B}                         & {\color[HTML]{333333} TruthfulQA(MC)}                       & {\color[HTML]{333333} 1.7342x}                                                                                    & {\color[HTML]{333333} 100}                                \\
\cellcolor[HTML]{FFFDFA}{\color[HTML]{333333} LLaMA-7B} & \cellcolor[HTML]{FFFDFA}{\color[HTML]{333333} Factor(Wiki)} & 1.8311x                                                                                                           & \cellcolor[HTML]{FFFDFA}{\color[HTML]{333333} 100}        \\
\rowcolor[HTML]{FFFDFA} 
{\color[HTML]{333333} LLaMA-7B}                         & {\color[HTML]{333333} StrategyQA}                           & {\color[HTML]{333333} 1.7542x}                                                                                    & {\color[HTML]{333333} 100}                                \\
\rowcolor[HTML]{FFFDFA} 
{\color[HTML]{333333} LLaMA-7B}                         & {\color[HTML]{333333} GSM8K}                                & {\color[HTML]{333333} 1.8883x}                                                                                    & {\color[HTML]{333333} 100}                                \\ \midrule
\rowcolor[HTML]{FFFDFA} 
{\color[HTML]{333333} LLaMA-13B}                        & {\color[HTML]{333333} TruthfulQA(MC)}                       & {\color[HTML]{333333} 1.8444x}                                                                                    & {\color[HTML]{333333} 100}                                \\
\rowcolor[HTML]{FFFDFA} 
{\color[HTML]{333333} LLaMA-13B}                        & {\color[HTML]{333333} Factor(Wiki)}                         & {\color[HTML]{333333} 1.9921x}                                                                                    & {\color[HTML]{333333} 100}                                \\
\rowcolor[HTML]{FFFDFA} 
{\color[HTML]{333333} LLaMA-13B}                        & {\color[HTML]{333333} StrategyQA}                           & {\color[HTML]{333333} 1.9929x}                                                                                    & {\color[HTML]{333333} 100}                                \\
\rowcolor[HTML]{FFFDFA} 
{\color[HTML]{333333} LLaMA-13B}                        & {\color[HTML]{333333} GSM8K}                                & {\color[HTML]{333333} 1.8292x}                                                                                    & {\color[HTML]{333333} 100}                                \\ \bottomrule
\end{tabular}}
\caption{Decoding latency analysis with 100\% of token extrapolated.}
\label{tab:latency_2}
\end{table}

\end{document}